# Classifier Learning with Supervised Marginal Likelihood

Petri Kontkanen, Petri Myllymäki, Henry Tirri
Complex Systems Computation Group (CoSCo)
P.O.Box 26, Department of Computer Science
FIN-00014 University of Helsinki, Finland
http://www.cs.Helsinki.FI/research/cosco/

## Abstract

It has been argued that in supervised classification tasks it may be more sensible to perform model selection with respect to a more focused model selection score, like the *supervised (conditional) marginal likelihood*, than with respect to the standard unsupervised marginal likelihood criterion. However, for most Bayesian network models, computing the supervised marginal likelihood score takes exponential time with respect to the amount of observed data. In this paper, we consider diagnostic Bayesian network classifiers where the significant model parameters represent conditional distributions for the class variable, given the values of the predictor variables, in which case the supervised marginal likelihood can be computed in linear time with respect to the data. As the number of model parameters grows in this case exponentially with respect to the number of predictors, we focus on simple diagnostic models where the number of relevant predictors is small, and suggest two approaches for applying this type of models in classification. The first approach is based on mixtures of simple diagnostic models, while in the second approach we apply the small predictor sets of the simple diagnostic models for augmenting the Naive Bayes classifier.

## 1 INTRODUCTION

Classification means the task of predicting the value of a discrete class variable, given the values of the other variables (predictors). In classifier learning to goal is to build accurate classifiers given a sample of classified instances, i.e., vectors consisting of the values of the predictors together with the corresponding value for the class variable. The accuracy of the classifier is measured according to some loss function — in this paper we focus on the 0/1-loss and logarithmic loss.

Classifier learning can be seen as a model selection process, where the goal is to search for the most accurate classifier in the chosen model family, the set of all models under consideration. Alternatively, the classifier can be built by using model averaging, in which case the final classifier is a weighted mixture of several individual models. In both approaches, one needs to define a criterion for model goodness that will be used in model selection or model averaging. As our goal was to build accurate classifiers from a sample of data, this criterion is naturally a function of the data.

In this paper we take the Bayesian approach, and assume our models to be probability distributions. More specifically, the model family consists of Bayesian network models that define probability distributions via a set of independence assumptions that can be conveniently expressed as a directed acyclic graph. In the Bayesian setting, the standard model selection (or model averaging) criterion is the posterior probability of the model, given the sample data. Assuming the uniform prior for the models, this is equivalent to using the marginal likelihood as model selection criterion. Nevertheless, recently it has been quite convincingly argued [10, 2] that although the joint marginal likelihood is in the Bayesian setting the typical choice for a model selection criterion, in supervised domains such as classification, it may be more sensible to use a more focused model selection criterion.

As discussed in [10], a natural supervised equivalent for the marginal likelihood is the supervised marginal likelihood, meaning the marginal of the conditional likelihood, where the likelihood is computed with respect to the values of the class variable in the sample data, given the value of the predictors. Unfortunately, for most Bayesian networks, computing this criterion is computationally infeasible. In this paper we consider a specific subclass of Bayesian networks, *diag-* *nostic Bayesian network classifiers*, where all the arcs connected to the class variable in the Bayesian network are arriving arcs, in contrast to *sampling* classifiers where all the arcs connected to the class variable are leaving arcs. A well-known example of the sampling Bayesian network classifier is the Naive Bayes classifier.

The advantage with the diagnostic classifiers is that in this case, computing the supervised marginal likelihood criterion is in principle easy. The caveat is that in this approach the number of parameters grows exponentially with the number of relevant predictors (the "parents" of the class variable). This problem is typically circumvented by assuming some simple, restricting form for the conditional distribution of the class variable, given its parents (see, e.g., [8] and the references therein).

In this paper we take a different approach, and restrict the number of relevant predictors to a relatively small number, in which case the supervised marginal likelihood can be computed efficiently. We then suggest two alternative approaches for exploiting this type of diagnostic models. In the first approach we form mixtures of several simple diagnostic classifiers, each having a small number of relevant predictors. The relevant predictor subsets can be either overlapping, or non-overlapping, in which case they form a partitioning of the predictor variables. In our second approach we apply these partitionings for constructing augmented Naive Bayes models, where the Naive Bayes independence assumption is relaxed by allowing dependencies within each predictor subset, and only assuming the subsets to be independent. This approach is in a sense "semi-supervised" as only the structure of the augmented Naive Bayes model is chosen according to supervised considerations, while the parameters of the model are treated according to the standard (unsupervised) Bayesian methodology, and not determined by supervised methods like the CEM algorithm [9]. Section 3 describes the two approaches in more detail.

In the experiments reported in Section 4, the first "mixture" approach yielded classifiers that were substantially more accurate on the average than the standard Naive Bayes classifier with respect to the conditional logarithmic loss, while with respect to the 0/1-loss their average predictive accuracy over several data sets was slightly worse. The augmented Naive Bayes classifiers obtained with the second approach were quite robust in the sense that they never showed much worse performance than the Naive Bayes classifier, while in some cases their accuracy was substantially better. Their average accuracy over several data sets was also significantly better than that of the vanilla Naive Bayes model with respect to both 0/1-loss and conditional logarithmic loss.

## 2 THE CLASSIFICATION PROBLEM

In classifier learning, the goal is to build accurate classifiers from a given *training data set* $\mathbf{D} = (\mathbf{x}^N, y^N)$, a matrix of $N$ vectors each consisting values of $n$ predictor variables $X_1, \ldots, X_n$, together with a value for a discrete class variable $Y$. For simplicity, in the sequel we assume the predictor variables $X_i$ to be discrete also.

In the probabilistic framework, the goal is to produce the classification predictive distribution $P(Y \mid X_1, \ldots, X_n)$. We can further distinguish two different approaches for estimating the classification distribution [3]: in the *diagnostic paradigm* one tries to estimate the predictive distribution directly, while in the *sampling paradigm* one estimates the distributions $P(X_1, \ldots, X_n \mid Y)$ and $P(Y)$, from which the desired classification predictive distribution can be computed by using the Bayes rule:

$$P(Y \mid X_1, \ldots, X_n) \propto P(X_1, \ldots, X_n \mid Y) P(Y). \quad (1)$$

Alternatively, one can also regard the joint distribution $P(X_1, \ldots, X_n, Y)$ as the basic building block in modeling, and the diagnostic paradigm and the sampling paradigm to be two alternative ways for factorizing this distribution.

The model families $\mathcal{F}$ we consider in this work consist of a finite number of probabilistic Bayesian network models $M$, $\mathcal{F} = \{M_1, \ldots, M_K\}$. A Bayesian network [13] is graphical representation for a set of independence assumptions between the domain variables[1]. The nodes of the directed acyclic graph correspond to variables, and the arcs represent the independence assumptions. One of the properties of these models is that the joint probability distribution can be factorized as follows:

$$P(X_1, \ldots, X_n, Y) = \prod_{i=1}^{n+1} P(X_i \mid \Pi_i), \quad (2)$$

where $\Pi_i$ denotes the *parents* (immediate predecessors in the graph) of variable $X_i$, and $X_{n+1}$ denotes the class variable $Y$. The parameters of a Bayesian network model determine the local conditional probability distributions $P(X_i \mid \Pi_i)$. This means that a Bayesian network graph $M$, together with the values for the model parameters $\Theta$, defines a joint probability distribution $P(X_1, \ldots, X_n, Y \mid M, \Theta)$ via (2).

---

[1] For an interactive tutorial on Bayesian network and causal modeling, and links to Bayesian network reference material, go to site http://www.b-course.cs.helsinki.fi.

Given a training set $\mathbf{D}$, with certain technical assumptions (see [7]), it is possible to compute the single model predictive distribution

$$P(X_1,\ldots,X_n,Y \mid M,\mathbf{D}) =$$
$$\int P(X_1,\ldots,X_n,Y \mid M,\Theta,\mathbf{D})P(\Theta \mid M,\mathbf{D})d\Theta. \quad (3)$$

If we now instead of using only a single model, average over all the Bayesian networks $M \in \mathcal{F}$, we get

$$P(X_1,\ldots,X_n,Y \mid \mathbf{D},\mathcal{F}) =$$
$$\sum_{M \in \mathcal{F}} P(X_1,\ldots,X_n,Y \mid M,\mathbf{D})P(M \mid \mathbf{D},\mathcal{F}), \quad (4)$$

where the first term was given in Equation (3). The second term is the posterior probability of the model $M$ after seeing the data $\mathbf{D}$. Intuitively, if one wants to choose a model from $\mathcal{F}$, it makes sense to select the model maximizing this posterior since that particular model has the highest overall weight in the sum (4). Assuming the prior $P(M \mid \mathcal{F})$ to be uniform, this is equivalent to choosing the model with the highest marginal likelihood $P(\mathbf{D} \mid M,\mathcal{F})$:

$$P(M \mid \mathbf{D},\mathcal{F}) \propto P(\mathbf{D} \mid M,\mathcal{F})P(M \mid \mathcal{F}). \quad (5)$$

As shown in [7], with the same technical assumptions mentioned above, the marginal likelihood can be computed in closed form.

In practice it is of course impossible to marginalize over all the possible Bayesian network models, the number of which is astronomical. For the same reason, finding the best model with respect to the marginal likelihood criterion is equally infeasible. Consequently, in practice we need to perform model search or model averaging over a subset of all Bayesian networks. Recently, it has been argued that in supervised domains it may be more sensible to not to use the marginal likelihood criterion for model selection/averaging, but to use a more focused criterion [5, 10]. This approach can be motivated by the following observation: the marginal likelihood criterion is related to predictive accuracy with respect to the joint probability distribution

$$P(\mathbf{D} \mid M,\mathcal{F}) = P(\mathbf{x}^N,y^N \mid M,\mathcal{F})$$
$$= P(y^N \mid \mathbf{x}^N,M,\mathcal{F})P(\mathbf{x}^N \mid M,\mathcal{F}). \quad (6)$$

If it happens that in $\mathcal{F}$ there are only a few good classifiers (models that are good with respect to the *supervised (conditional) marginal likelihood*)

$$P(y^N \mid \mathbf{x}^N,M,\mathcal{F}) =$$
$$\int P(y^N \mid \mathbf{x}^N,M,\mathcal{F},\Theta)P(\Theta \mid \mathbf{x}^N,M,\mathcal{F})d\Theta, \quad (7)$$

then the unsupervised marginal likelihood criterion tends to favor models that model well the predictor marginal likelihood $P(\mathbf{x}^N \mid M,\mathcal{F})$, which however is irrelevant with respect to the classification task. For this reason, it has been suggested [5, 10] that in practice one should resort to some supervised model selection criterion, like the *supervised marginal likelihood* (also called the *class sequential criterion* [8]). The recent empirical results reported in [2, 12, 10] support this line of reasoning.

## 3 BAYESIAN NETWORK CLASSIFIERS

When determining the model family used, the subset $\mathcal{F}$, what type of Bayesian network models should one consider? As discussed above, the joint distribution represented by a Bayesian network model can be used for producing classifiers either according to the sampling paradigm or according to the diagnostic paradigm. In the former case all the arcs connected to the class variable node would be leaving arcs, in the latter case arriving arcs. Actually, the Bayesian network formalism allows also a hybrid approach where the class node is connected to the rest of the network both via leaving arcs and arriving arcs. A trivial observation is that in classification, only the nodes belonging to the so called *Markov blanket*[2] of the class node are relevant, which follows from the properties of the Bayesian network models. In the following we focus on the pure diagnostic and sampling approaches.

An example of the sampling Bayesian network classifier is the Naive Bayes classifier, a Bayesian network with one arc from the class node to each of the predictor nodes (see Figure 1). This graph structure represents the assumption that the predictors are independent of each other, given the value of the class variable. This assumption sounds of course very naive, but rather surprisingly, the Naive Bayes classifier is in fact in many real-world cases the state-of-the-art classifier, as for example, its success in prediction competitions like the KDD Cup and the CoIL competition illustrate.

Nevertheless, the structural simplicity of the Naive Bayes classifier calls for developing better alternatives. In [5] this problem was approached by augmenting the Naive Bayes model by adding arcs between the predictor variables. In the empirical tests reported, some improvement was obtained with the tree-augmented approach, but the difference was not dramatic. Similar results were reported in [10], where the Naive Bayes

---
[2]The Markov blanket of a node consists of the parents of the node, the children (immediate successors) of the node, and the parents of the children.

model was augmented by feature selection.

The problem with the augmented Naive Bayes approach is that, as discussed in [5, 10], the supervised marginal likelihood criterion is computationally infeasible. In [12], the *prequential model selection criterion* showed some promise, but nevertheless, it seems that the interesting supervised model selection criteria are difficult to compute for the sampling Bayesian network classifiers, i.e., the Naive Bayes model and the augmented versions of it.

In the following we take a radically different viewpoint, and look at the diagnostic Bayesian network classifiers. In this case, with the assumptions mentioned above, computing the supervised marginal likelihood (7) is in principle easy:

$$P(y^N \mid \mathbf{x}^N, M, \mathcal{F}) = \prod_{j=1}^{q} \frac{\Gamma(N'_j)}{\Gamma(N'_j + N_j)} \prod_{k=1}^{r} \frac{\Gamma(N'_{jk} + N_{jk})}{\Gamma(N'_{jk})}, \quad (8)$$

where $\Gamma$ denotes the gamma function, $q$ is the number of value configurations for the predictors $X_1, \ldots, X_n$, $r$ is the number of values of the class variable $Y$, $N_{jk}$ are the sufficient statistics (the number of cases where the predictor values are in configuration $j$ when the class variable has value $k$), and $N_j = \sum_{k=1}^{r} N_{jk}$. The constants $N'_{jk}$ are the *hyperparameters* determining the parameter prior distribution $P(\Theta \mid M, \mathcal{F})$.

Note that in contrast to the work in [8], we do not assume any specific simplifying form for the conditional distribution $P(y^N \mid \mathbf{x}^N, M, \mathcal{F})$, but use the general multinomial distribution. It now follows of course that the number of parameters required is exponential with respect to the number of predictors $X_i$, or more precisely, if not all the predictors are connected to the class variable, with respect to the number of arcs arriving to the class variable. Nevertheless, as computing the supervised marginal likelihood is feasible in this framework, we suggest two approaches for exploiting this observations in building practically feasible classifiers. The first approach is based on a mixture of several simple diagnostic Bayesian network classifiers, while the second approach is based on a modification of the Naive Bayes classifier.

### 3.1  Mixtures of diagnostic Bayesian network classifiers

As can be observed later, for our purposes it is convenient to consider diagnostic models where the predictors are linked to each other via a fully connected network[3]. In this framework, we can identify the individual models by the arcs arriving to the class variable, by listing the corresponding *relevant predictor variables*. The number of parameters is now exponential with respect to the number of the relevant predictors. An obvious way to cope with the exponentiality is to restrict the number of arcs in the network to some relatively small number. For example, we might define our model family $\mathcal{F}$ to consist of all the diagnostic models where the relevant predictors consist of exactly 3 variables. Figure 1 shows two such models with overlapping relevant predictor subsets of size 3.

Let us now in this framework have a look at the classification predictive probability for the class variable $y$, given a classification query $\mathbf{x}$, and the training data $\mathbf{D} = (\mathbf{x}^N, y^N)$. By averaging over all the models $M$ in $\mathcal{F}$, we get

$$\begin{aligned}
P(y|\mathbf{x}, \mathbf{x}^N, y^N, \mathcal{F}) &= \\
\sum_{M \in \mathcal{F}} & P(y|\mathbf{x}, \mathbf{x}^N, y^N, \mathcal{F}, M) P(M|\mathbf{x}, \mathbf{x}^N, y^N, \mathcal{F}) = \\
\sum_{M \in \mathcal{F}} & \Big( P(y|\mathbf{x}, \mathbf{x}^N, y^N, \mathcal{F}, M) P(y^N|\mathbf{x}^N, M, \mathcal{F}) \\
& \cdot P(\mathbf{x}^N|M, \mathcal{F}) P(M|\mathcal{F}) \frac{P(\mathbf{x}|\mathbf{x}^N, y^N, M, \mathcal{F})}{P(\mathbf{x}, \mathbf{x}^N, y^N, \mathcal{F})} \Big). \quad (9)
\end{aligned}$$

As we above assumed that the predictors are connected to each other via a fully connected network in all the models $M \in \mathcal{F}$, $P(\mathbf{x}^N \mid M, \mathcal{F})$ is a constant and can be ignored. Moreover, assuming the uniform prior over the models $M$ in $\mathcal{F}$, and leaving out terms that do not depend on $y$, we get

$$P(y|\mathbf{x}, \mathbf{x}^N, y^N, \mathcal{F}) = \sum_{M \in \mathcal{F}} P(y|\mathbf{x}, \mathbf{x}^N, y^N, \mathcal{F}, M) P(y^N|\mathbf{x}^N, M, \mathcal{F}). \quad (10)$$

Consequently, the result is a finite mixture of several diagnostic Bayesian network classifiers, where the individual predictions made by the models $M \in \mathcal{F}$ are weighted by the supervised marginal likelihood (8).

In the above framework, we can even move one step further in the abstraction level and ask how to choose the model family $\mathcal{F}$ — how do we determine the relevant predictor variable subsets? One solution is to fix the size of the relevant subsets to some constant, say 3, and define $\mathcal{F}$ to consist of all the possible subsets with 3 variables. This approach is of course possible in practice only if we restrict the size of the subsets to a relative low number. An alternative approach is to require the subsets to be non-overlapping, in which case the relevant subsets form a partitioning of the predictor variables, and the number of resulting mixture components is bounded above by the total number of predictors.

---

[3] Actually, the structure of the predictor subnetwork is irrelevant, as long as it is the same for all the models $M$.

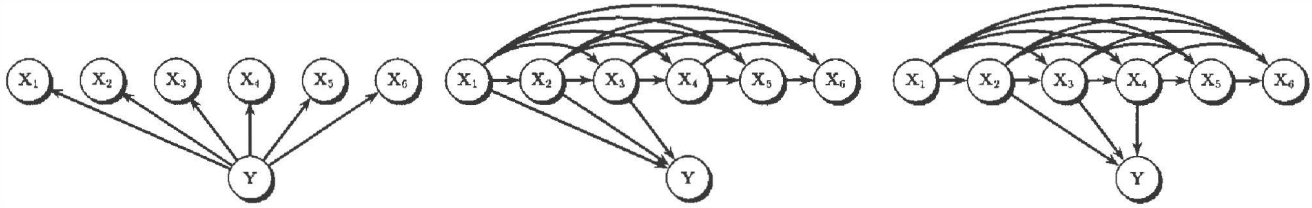

Figure 1: The "vanilla" Naive Bayes model (on the left) and two examples of diagnostic Bayesian network classifiers with overlapping relevant predictor subsets of size 3.

Let $\mathcal{F}_1$ and $\mathcal{F}_2$ now be two such model families, each corresponding to a different partitioning of the predictors. According to the discussion above, we should evaluate the model families according to the meta level form of the supervised marginal likelihood:

$$P(y^N \mid \mathbf{x}^N, \mathcal{F}) = \sum_{M \in \mathcal{F}} P(y^N \mid \mathbf{x}^N, M, \mathcal{F}) P(M \mid \mathbf{x}^N, \mathcal{F}). \quad (11)$$

As we assumed the predictors to be fully connected to each other, it is easy to see that $P(M \mid \mathbf{x}^N, \mathcal{F})$ is a constant and this corresponds to computing the average of the supervised marginal likelihoods with respect to models $M$ in $\mathcal{F}$.

### 3.2 Augmenting the Naive Bayes classifier

In the partitioning approach discussed above, we presented a method for finding such subsets of predictor variables that the mixture of the corresponding diagnostic Bayesian network classifiers is a good model with respect to the criterion (11). Intuitively, one would expect the emerging predictor subsets to be such that the predictors in the same subset are highly dependent of each other.

We can now exploit this observation in the Naive Bayes classifier framework, where dependencies between the predictors violate the basic independence assumption. One way to relax the independence assumption is to merge highly relevant predictors together to a meta-attribute, or equivalently, connect them via a fully connected subnetwork. For doing this, we obviously need a method that finds subsets of attributes that are highly dependent of each other. This is exactly what we argued is the result of the partitioning diagnostic Bayesian network approach. Consequently, we propose an augmented Naive Bayes classifier, where the attributes are merged in such a way that the criterion (11) is maximized.

To illustrate the idea further, let us consider a simple classification problem with 6 predictors $X_1, \ldots, X_6$, and let us assume that the best partitioning found by using the criterion (11) is $\{\{X_1, X_2, X_3\}, \{X_4, X_5\}, \{X_6\}\}$. Figure 2 shows the corresponding diagnostic classifiers, over which we can form a mixture as discussed in the previous subsection. The corresponding augmented Naive Bayes model can be found in the same figure.

## 4 EMPIRICAL RESULTS

The classification models discussed in Section 3 were empirically tested by using 15 classification data sets from the UCI data repository [1]. In this set of experiments, the data sets were preprocessed by discretizing the continuous variables by using the simple equal-frequency discretization with 3 bins. Consequently, from the point of view of this work, all the data sets were discrete.

In the sequel, the following notation is used for the different classifiers used:

**OMi**   A mixture of diagnostic Bayesian network classifiers, where the relevant predictor subsets are of size $i$, and they can be overlapping. The mixture goes exhaustively over all the subsets. For computational reasons, $i$ must of course be relatively small in practice.

**PM**   A mixture of diagnostic Bayesian network classifiers, where the relevant predictor subsets form a partitioning of the predictors. Among all the possible partitionings, the best according to the meta learning criterion (11) is chosen. The number of components in the resulting mixture (the number of predictor subsets in the partitioning), is determined automatically by the learning criterion. In this set of experiments, the search algorithm used for finding good partitionings was a simple stochastic greedy clustering method.

**ANB**   An augmented Naive Bayes model where the partitioning found by PM is used to form cliques of predictors, where the clique subnetwork are fully connected, but there are no arcs between the cliques (see Figure 2).

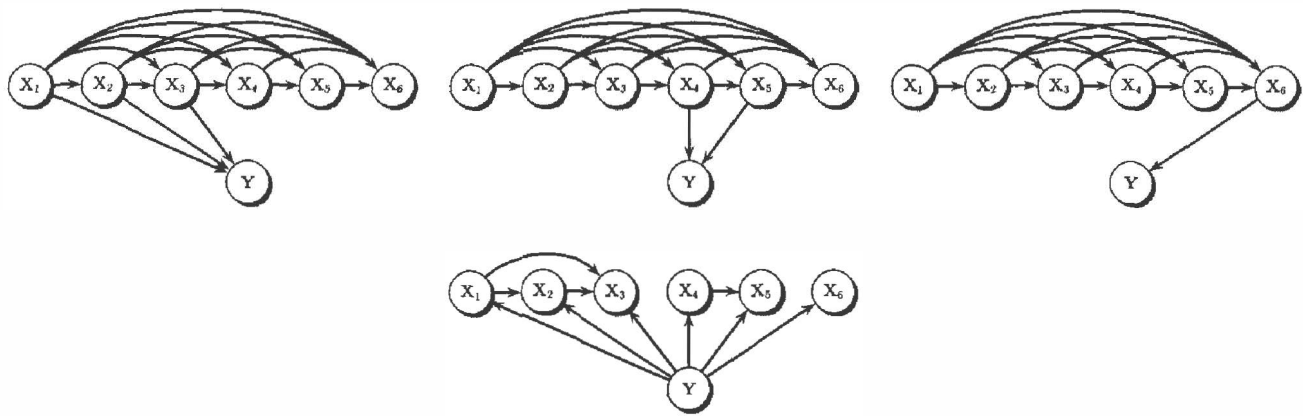

Figure 2: On top: the three diagnostic classifiers corresponding to the predictor partitioning $\{\{X_1, X_2, X_3\}, \{X_4, X_5\}, \{X_6\}\}$. A mixture over these models produces the corresponding PM classifier. Below: the corresponding ANB classifier obtained with the same partitioning.

With each data set, the classifiers were empirically validated by splitting the data randomly into a training set with 75% of the data, and a test with 25% of the data. The classification accuracy in the test set was measured both according to the 0/1-loss and the conditional logarithmic loss, and the results were scaled with respect to the accuracy obtained by the "vanilla" Naive Bayes model. The reported results are averages over 50 trials (50 training set–test set splits) with each data set.

A more commonly used empirical method for validating classification algorithms is the crossvalidation method. The reason for us not to use this method is that according to our experience (see e.g. [11]), the result obtained with the K-fold crossvalidation depends heavily on the way one partitions the data into the K folds (see also [4]). Leave-one-out crossvalidation does not have this drawback, but we have observed some anomalies caused by the smallness of the test set (which is of size 1). We could of course average the K-fold crossvalidation over several random fold partitionings, but that brings us essentially back to the training set-test set setup described above.

The results for all the data sets used can be found in Figures 3 and 4. Averages over the data sets are reported in Figure 5.

From the empirical results, we can make the following observations. In the 0/1-loss case, only the ANB classifier was able to beat the vanilla Naive Bayes classifier on the average. This can be explained by noting that the diagnostic mixtures were based on the Bayesian model averaging framework, which is designed for the conditional logarithmic loss case, while the ANB classifier is more of an ad hoc approach in the spirit of the Naive Bayes classifier. This line of reasoning is supported by the log-loss results, where the diagnostic models outperform the Naive Bayes classifier clearly (with the exception of the simplistic OM1 model), and they in this case perform better than the heuristic ANB classifier.

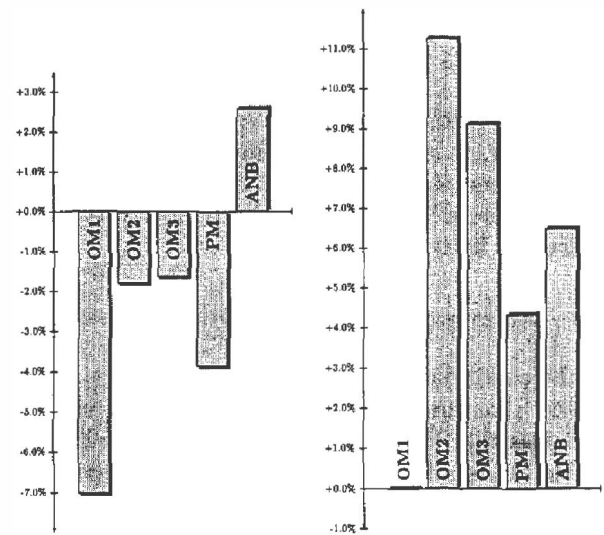

Figure 5: Average 0/1-loss (left) and conditional logarithmic loss (right) prediction gains over the vanilla Naive Bayes classifier, averaged over all the data sets.

The ANB model seems to be a very robust classifier, which in the 0/1-loss case never collapses with respect to the Naive Bayes classifier, and in some cases perform substantially better. From Figure 4 we can see that the ANB classifier was also the most robust classifier in the log-loss case, although the overall average performance was in this case better with the diagnostic classifiers.

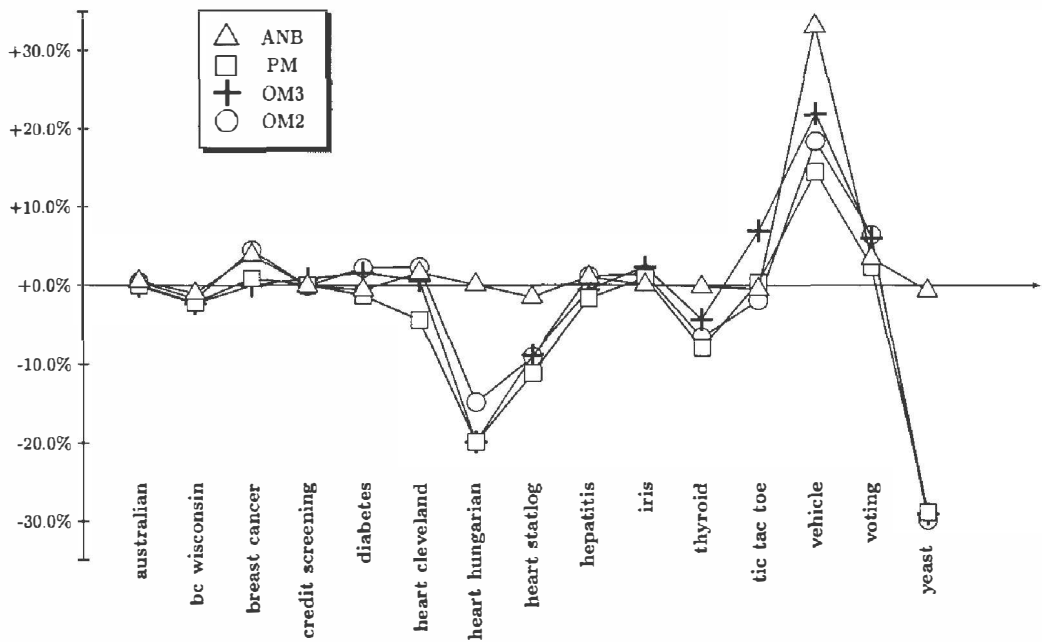

Figure 3: Results with different data sets with respect to the 0/1-loss. The results are scaled with respect to the accuracy obtained with the vanilla Naive Bayes classifier.

## 5　CONCLUSIONS AND FUTURE WORK

We have studied diagnostic Bayesian network classifiers where the relevant model parameters are directly related to the classification task. We suggested two possible approaches for applying this framework in classifier learning. The first approach was based on model averaging: the resulting classifier is a weighted mixture of the small diagnostic models found, where the weights are the corresponding supervised marginal likelihood values. In the second approach we did not apply the diagnostic models directly in classification, but only used the corresponding predictor subsets for augmenting the Naive Bayes classifier. The basic idea was that when learning diagnostic classifiers, the resulting predictor subsets represent interesting variable combinations. One can now relax the basic Naive Bayes independence assumption by assuming independence between the variable subsets, not within a subset. An obvious extension of this work would be to allow some dependencies between the fully connected subnetworks.

The empirical results showed that with respect to the logarithmic loss, the diagnostic OMi and PM models performed on the average significantly better than the Naive Bayes classifier. The ANB model was the most robust classifier, although the overall average was not with this loss function as good as with the diagnostic approached.

The mixtures of diagnostic classifiers were designed for the logarithmic loss case, which explains why their performance with respect to the 0/1-loss was not very good. It would be an interesting research problem to modify the model averaging approach for some other utility function, for example by adopting the general *entropification* framework described in [6]. We would also like to mention that in addition to the results obtained with the UCI data sets, we have experimented with these methods in a real-world medical problem domain. In this case, the diagnostic approaches yield much better results than ANB or NB. We are currently investigating the reason for this observation by analyzing the differences between the UCI data sets and our real-world data sets.

With respect to the 0/1-loss, the ANB classifier was a clear winner in the tests reported here. All in all, all our empirical results suggest that the ANB model is a very robust classifier that never gives much worse result than the Naive Bayes classifier, while it in some cases performs significantly better. There are several ways to improve the ANB setting used in this work (for instance, one could use a more elaborate search algorithm than the simple stochastic greedy method used here), so the ANB approach seems to offer a promising framework for developing accurate classifiers.

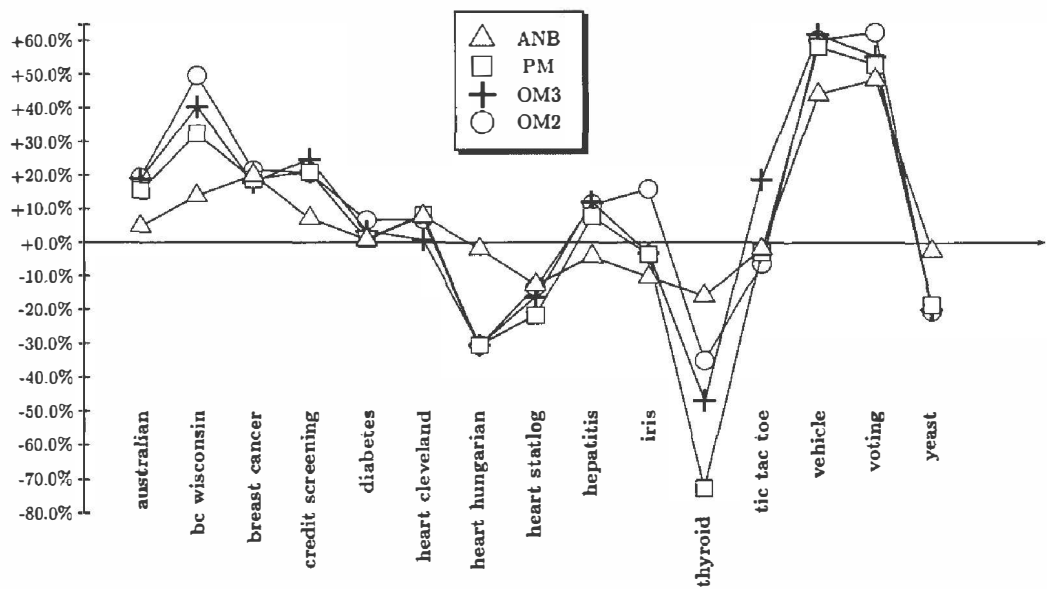

Figure 4: Results with different data sets with respect to the conditional logarithmic loss. The results are scaled with respect to the accuracy obtained with the vanilla Naive Bayes classifier.

## Acknowledgements

This research has been supported by the National Technology Agency, and the Academy of Finland.